\documentclass[11pt]{article}
\usepackage[preprint]{acl}
\usepackage{times}

\usepackage{latexsym}
\usepackage[T1]{fontenc}
\usepackage[utf8]{inputenc}
\usepackage{amsmath,amssymb}
\usepackage{microtype}
\usepackage{booktabs}
\usepackage{graphicx}
\usepackage{listings}
\usepackage{xcolor}
\newcommand{\difftext}[1]{#1}
\usepackage{url}

\title{Memory-Aware Reward Modeling:\\Learning Human Preferences for LLM Memory Use}
\author{
Yijiang River Dong$^{1,*}$ \quad Zexue He$^{2,*}$ \quad Cheng Yu$^{3,*}$ \quad Tiancheng Hu$^1$ \quad Michael J Ryan$^2$ \\
\textbf{Jan Batzner$^3$ \quad Zheng Hui$^1$ \quad Shaolong Wu$^4$ \quad Orestis Papakyriakopoulos$^3$} \\
\textbf{Alex Pentland$^2$ \quad Nigel Collier$^{1,\dagger}$ \quad Jiaxin Pei$^{5,\dagger}$} \\[0.5em]
{\normalfont\normalsize $^1$University of Cambridge \quad $^2$Stanford University \quad $^3$Technical University of Munich} \\
{\normalfont\normalsize $^4$Harvard University \quad $^5$University of Texas at Austin} \\[0.3em]
{\normalfont\small $^*$Equal contribution. \quad $^{\dagger}$Equal advising.}
}

\begin{document}
\maketitle
\begin{abstract}

Memory enables language model assistants to maintain continuity across sessions and generate personalized responses, but knowing more about a user does not necessarily produce responses they prefer. Existing evaluations assess memory capabilities without directly establishing whether memory use benefits the user. To study this gap, we collect approximately 1,000 pairwise judgments from 87 participants using their own ChatGPT queries and personal memories. We find that memory-conditioned responses are not consistently preferred over responses generated without memory and in some cases, users strongly prefer the latter. Automatic evaluators also show limited agreement with users and favor memory-conditioned responses substantially more often than users do. Motivated by these gaps, we introduce a memory-aware reward model that learns from user judgments, achieving 72.8\% preference prediction accuracy and outperforming the existing baselines. Our findings motivate evaluation and optimization objectives that align memory use with the preferences of the people whose information is being used. Our code is available at 
\url{https://github.com/dong-river/memory-aware-RM}.

\end{abstract}
\begin{figure*}[!t]
\centering
\includegraphics[width=\textwidth]{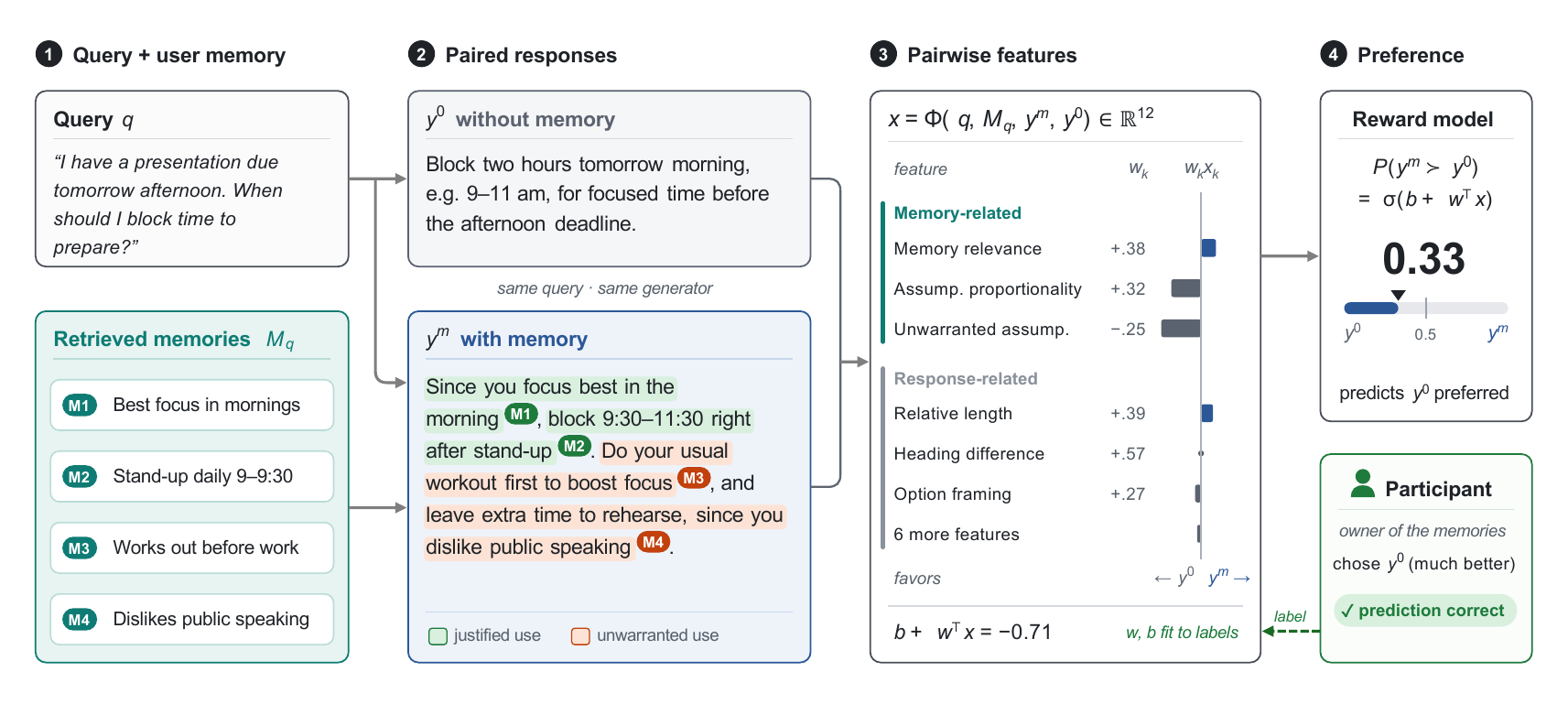}
\caption{\textbf{Overview of memory-aware preference learning.}
Given a query and retrieved user memories, the same generator
produces paired responses with ($y^m$) and without ($y^0$)
memory conditioning. A rubric evaluator assesses memory-related
and general response characteristics, yielding pairwise features
$x$. Feature weights are learned from annotators' pairwise
preferences, allowing the model to predict which response is
preferred. Importantly, annotators evaluate responses generated
using their own memories. The example and feature contributions
are illustrative; Section~\ref{sec:method} describes the full
model.}
\label{fig:overview}
\end{figure*}

\section{Introduction}
\label{sec:intro}
Memory has emerged as a key component of conversational language models, enabling them to retain and reuse information beyond their context length and across multiple conversation sessions. Existing work has explored mechanisms for storing past experiences, managing information across context windows, and retrieving knowledge to support subsequent reasoning and action \citep{zhong2023memorybank,packer2023memgpt}. These capabilities support continuity across interactions and tasks. In persistent assistants, memory allows models to employ information about the user including their persona, their expressed preferences, and past requests to generate more personalized and contextually appropriate responses.

Much of the progress in long-term memory has been measured by and optimized for benchmarks like LongMemEval \citep{wu2025longmemeval}, Locomo \citep{maharana2024locomo}, and PersonaMem \citep{jiang2025personamem} that evaluate whether models can compress and recover information from past interactions to the current session. However, accurate recall and successful preference inference do not ensure appropriate memory usage. Consider a user who previously requested explanations tailored to their profession and now asks for a general overview of a topic. Reusing that preference may appear relevant and personalized, yet narrow the answer in a way the current request does not warrant. Recent studies document such failures through over-personalization and inappropriate memory use \citep{hu_op-bench_2026,feng_rpeval_2026,pulipaka_persistbench_2026, zhang2026re}. These errors cannot be characterized solely as failures of memory accuracy or retrieval. They mainly concern how available information should be used for LLM response.

To understand whether memory-based personalization produces responses that users prefer, we collect \textbf{approximately 1,000 response-pair judgments from 87 participants}. Participants construct personal memory profiles from their own ChatGPT interactions and recent experiences, then evaluate paired responses to queries drawn from their interaction histories. Each pair uses the same query and generator, with one response receiving retrieved memories and the other generated without them. Crucially, participants judge responses grounded in their memories, allowing us to examine the value of personalization from user's perspective.

Our findings reveal a gap between \textbf{personalizing a response using the user's memory and producing a response that user prefers}. Memory-conditioned responses do not consistently outperform responses generated without memory, and personalization sometimes backfires as users strongly prefer the response generated without their personal context. These findings challenge the assumption that incorporating relevant personal information necessarily improves the user experience and motivate  new evaluation and optimization objective for memory-augmented language models.

We next examine whether automatic evaluators capture these preferences. When given access to the user's memories alongside the query and candidate responses, a direct LLM judge and pretrained reward models show limited agreement with users, with prediction accuracy close to the majority baseline. They also select memory-conditioned responses substantially more often than users do. This discrepancy identifies a second gap: \textbf{automatic evaluation of memory-conditioned responses do not reliably reflect the user's preferences}.

To examine these gaps, we explore an initial
interpretable framework for analyzing how
memory use relates to user preferences,  In
conventional reward modeling, a model learns \(R_\theta(q,y)\) from human comparisons of responses to a user query \(q\) \citep{ouyang2022instruct}. We introduce \textbf{Memory-Aware Reward Model} that make user memory \(M\) explicit in \(R_\theta(q,y,M)\), learning preferences over both response quality and the appropriateness of memory use. Crucially, our supervision comes from real users judging responses grounded in their own memories, providing direct evidence of when and how they want remembered information to influence an answer. Applying rubric-based reward modeling to this problem, we achieve 72.8\% accuracy, compared with 59.6\% for a direct zero-shot LLM judge and 58.8\% mean accu-racy for pretrained rewards. 

Our contributions are threefold (1) a dataset of response preferences grounded in participants' own queries and memories, enabling direct study of how users experience memory-based personalization. (2) identification of two evaluation gaps: personalization does not consistently yield user preferred responses, and the existing automatic evaluators do not capture users' choices. (3) a memory-aware preference model that provides a interpretable proxy for human preference of memory usage that outperform the existing baselines. Together, these contributions establish human preference as a necessary point of reference for assessing the value of memory in personalized assistants.
\section{Related Work}
\label{sec:related}
\paragraph{Long-term LLM memory.}
Long-term memory systems enable language model assistants to accumulate and reuse information beyond a single context window. Existing approaches span external retrieval \citep{zhong2023memorybank,packer2023memgpt,chhikara2025mem0}, graph-based memory stores \citep{xu2025amem,jiang_magma_2026}, learned policies for memory management \citep{wang_mem-_2025, yan2026memory}, and parametric memory \citep{wang2024memoryllm, lei2026delta}. A growing set of benchmarks evaluates these capabilities through recall and multi-session reasoning \citep{maharana2024locomo,wu2025longmemeval,tan_membench_2025}, dynamic user profiling and personalized responses \citep{jiang2025personamem, jiang2025personamem-v2}, and tracking evolving user or world states \citep{liu_perma_2026,li_horizonbench_2026,fan_can_2026}. These evaluations primarily measure whether information can be retained, recovered, and updated, leaving less clear how an assistant should behave once relevant memories are available in context. Recent work moves beyond recall to examine whether past chat history supports downstream actions \citep{he_memoryarena_2026,shen_mem2actbench_2026,feng_know_2026}. Our work extends this focus by analyzing user preference and learning rubric-based proxies for memory usage grounded in real human preferences.

\paragraph{Personalization and Over-personalization.} LLM personalization has traditionally represented users through profiles, personas, demographic or preference attributes, and collections of prior interactions, which are then used to condition generation or predict individual preferences \citep{salemi2024lamp,kirk2024prism,dong2024personalizedjudge,ryan2025synthesizeme}. However, incorporating personal information does not always improve a response: recent work identifies \textbf{over-personalization} as a failure of using user information when it can feel unnecessary, distracing, sycophantic, or even privacy-intrusive \citep{hu_op-bench_2026,feng_rpeval_2026}. OP-Bench evaluates irrelevant personalization, repetition, and sycophancy, while PersistBench examines cross-domain memory leakage and memory-induced sycophancy \citep{hu_op-bench_2026,pulipaka_persistbench_2026}.
We instead treat appropriate memory use as a preference learning problem. Rather than assuming a fixed notion of when personalization is desirable, we collect response-level preferences from users over controlled variations of memory-conditioned responses and use these judgments to recover  signals for \emph{how} user memory should shape a response.

\paragraph{RLHF and post-training.}
Post-training aligns language models with human preferences through learned reward signals or direct preference optimization \citep{ouyang2022instruct,bai2022training,rafailov2023dpo}. Standard formulations assess candidate responses to a single prompt, whereas recent work extends preference modeling to multi-turn conversations and agent trajectories \citep{shani2024multiturn,shi2024direct}, or conditions judgments on user-specific summaries and learned user representations \citep{nam2026learning,shenfeld2025language,li2024personalized}. Beyond scaler reward model, rubric-based reward model raises for its interpretability, sample-efficiency, performance \citep{gunjal2026rubrics,liu2025inference,liu-etal-2026-openrubrics,ryan2025autometrics}. We extend preference-based alignment to cross-session memory, learning a rubric-based reward model that conditions on user memory and learns whether and how it should influnce the current response.

\section{Data Collection}
\label{sec:data}
We recruit participants from Prolific for a roughly 30-minute annotation task, paid at 14 USD per hour, and collect approximately 1,000 response-pair judgments (1,047 in total: 972 preference judgments and 75 test--retest repeats). We exclude the repeats from training and evaluation and filter the annotations for quality, leaving \textbf{821 judgments from 87 participants}, of which 673 are strict preferences and 148 are ties. Each judgment records an annotator's preference between two responses to the same query in the context of their personal memories. Evaluation separates participants across training, validation, and test within each split (Section~\ref{sec:experimental-setup}).

\subsection{Personal Context Collection}
Participants accessed the study through Prolific and constructed an editable memory profile by running an extraction prompt in their own ChatGPT account. The prompt draws on previous ChatGPT sessions and existing memories to (1) produce user profile such as age, occupation, location, etc. (2) infer user preferences around 10 topics including entertainment, lifestyle, communication, etc. (3) extract 100 their recent queries, excluding those dependent on unavailable attachments or prior context. Participants then entered specific events from the preceding month through typing or editable voice transcription, with a target of 30 events. 

Note that we carefully avoid uploading any personal data to our servers: full conversation histories and attachments were not uploaded and we explicitly instruct the model and participants to avoid including sensitive information. Appendix~\ref{app:extraction} reproduces the complete prompt, and Figures~\ref{fig:persona-ui}--\ref{fig:events-ui} show the collection interfaces. PII removal and de-identification procedures are described in Appendix~\ref{app:pii}.

\subsection{Response Pair Construction}
For each participant, we construct a candidate pool of 100 recent queries. For each query $q$, we retrieve a context $M_q$ consisting of nine memory entries using \emph{text-embedding-3-small} and randomly select a response generator from GPT-5.2, Gemini-3.1-Pro, and Claude-Opus-5. The selected model generates two responses in separate calls: one conditioned on the retrieved context $M_q$ and one without memory. The two responses thus share the same query and generator but differ in access to personal context.

For memory-conditioned responses, we vary the instructions governing memory use across three prompting regimes: \emph{conservative}, \emph{ordinary}, and \emph{aggressive}. The ordinary prompt instructs the model to incorporate memories where relevant and appropriate. The conservative prompt imposes a stricter criterion, instructing the model to use memories only when they would meaningfully improve the answer to the current query. The aggressive prompt encourages proactive incorporation of personal context whenever possible. Appendix~\ref{app:generation-prompts} provides the complete prompt templates.

We also screen the 100 candidate pairs to select 15 that have substantial, memory-grounded differences in responses for human annotation. An LLM evaluator receives each query, its retrieved memories, response without memory, response with memory (using ordinary prompt) and assesses the magnitude of their substantive differences, the extent to which memory supports those differences, and the presence of unsupported changes. We rank pairs by prioritizing substantial, memory-grounded differences and retain the 15 highest-ranked pairs.

\subsection{Response Preference Annotation}
Participants viewed the query and Response A and Response B side by side. They then select the answer they would prefer to receive on a five-point scale: A much better, A slightly better, about the same, B slightly better, or B much better. Optional notes allowed them to explain their choice. Instructions emphasized overall preference without prescribing greater personalization as desirable (Appendix~\ref{app:preference}). Appendix~\ref{app:examples} presents two complete examples with contrasting preferences.

Finally, we exclude incomplete response pairs, recorded attention-check failures, or other abnormal annotator behaviors, removing 151 of the 972 primary judgments and retaining 821 from 87 participants.

\subsection{Annotator Consistency}
\label{sec:retest-results}
Ten participants returned three days later to repeat judgments for identical queries and stored response pairs, with prior choices hidden. We measure directional agreement: whether the same response is preferred in both sessions, regardless of changes in preference strength. Overall agreement was 68.0\% across 75 comparisons. Initially \emph{much} preferences were more consistent than \emph{slightly} preferences (73.9\% versus 58.6\%; Table~\ref{tab:retest}). Due to the low consistency of the \emph{slightly better subset}, we only use the \emph{much better subset} that contains more consistent and clear preference signal for testing models in \ref{sec: results}.

\begin{table}[ht]
\centering
\setlength{\tabcolsep}{4pt}
\begin{tabular}{@{}lcr@{}}
\toprule
Original strength & Same side on repeat & Rate\\
\midrule
Slightly & 17/29 & \textbf{58.6\%}\\
Much & 34/46 & \textbf{73.9\%}\\
\midrule
Overall & 51/75 & \textbf{68.0\%}\\
\bottomrule
\end{tabular}
\caption{Test--retest agreement among 10 participants after three days. ``Same side'' counts comparisons in which the same response is preferred in both sessions; groups are defined by the original preference strength.}
\label{tab:retest}
\end{table}

\section{Method}
\label{sec:method}
\subsection{Preliminaries}
\label{sec:prelim}
\paragraph{Reward models.}
Reward modeling turns human judgments into a scoring function for selecting or optimizing model outputs. Given query $q$ and response $y$, a scalar reward model $R_\theta(q,y)$ represents response quality through a learned utility. Pairwise preferences are commonly modeled as
\begin{equation}
P_\theta(y_a\succ y_b\mid q)
=\sigma\!\left(R_\theta(q,y_a)-R_\theta(q,y_b)\right),
\label{eq:reward}
\end{equation}
where $\sigma$ is the sigmoid function. Training increases the reward of preferred responses relative to rejected ones \citep{bradley1952rank,schulman2017proximal}.

\paragraph{Rubric-based reward models.}
Rubrics express response quality through explicit criteria, such as correctness, relevance, or respect for user constraints. An evaluator assesses these criteria, and an aggregation rule converts its judgments into a reward or preference. This separates specifying desirable behavior from assessing whether a response exhibits it \citep{gunjal2026rubrics,liu2025inference}. A simple scalar formulation is $R_{\mathbf w}(q,y)=\mathbf w^\top\boldsymbol\phi(q,y)$, where $\boldsymbol\phi$ contains feature scores that is typically assigned by an external evaluator (e.g., LLM Judge) and $\mathbf w$ is the weight vector of features.

\subsection{Memory-Aware Preference Modeling}
\label{sec:formulation}
Each example contains a query $q_i$, retrieved memories $M_i$, a memory-conditioned response $y_i^m$, a response generated without memory $y_i^0$, and the annotator's preference. Our objective is to predict which response the annotator prefers. We represent each response comparison using interpretable features:
\begin{equation}
\mathbf x_i=\Phi(q_i,M_i,y_i^m,y_i^0)\in\mathbb R^{12}.
\end{equation}

\textbf{Feature initialization.} Preference annotations are used to induce candidate evaluation criteria. Each induction pass receives ten preference-labeled quadruplets from training set: the query, the retrieved memory set, both candidate responses, and which response the annotator chose. Then LLM then generates ten rubric dimensions that explain the difference between the two responses, such as response quality, formatting, or memory usage, etc.  We run five independent passes, yielding roughly fifty raw proposals, many of which describe the same property in different words. A separate consolidation pass merges near-duplicates into a canonical rubric and labels each dimension memory-dependent if scoring it requires the memory contents, and memory-agnostic otherwise. Fixing that label at induction time means the decomposition in §6 partitions the features by definition rather than by our later judgement.

For example, in one experiment, the procedure above gives us 12 features, including three memory-related features (memory relevance, assumption proportionality, and unwarranted personal assumptions) and nine response-quality features describing response quality, framing, and form. Table~\ref{tab:all-features} defines all twelve features. Note that rubric induction and consolidation are performed independently within each outer cross-validation fold, using only that fold’s training participants and their preference labels. The resulting rubric is held fixed within the fold for feature extraction and regression, and the learned predictor is evaluated on the held-out participants.

\textbf{Feature regression.} Following \citet{ryan2025autometrics,li2026harmrecovery}, we learn weights that aggregate the
features into the annotator's pairwise preference. Each entry of $\mathbf x_i$ compares $y_i^m$ with $y_i^0$ on one feature
(e.g., heading count, or a judge's memory-relevance score), signed so that larger values favor $y_i^m$. Let $z_i=1$ if the
annotator prefers $y_i^m$ and $z_i=0$ otherwise, and let $\widetilde{\mathbf x}_i$ be $\mathbf x_i$ standardized on the training
set. We fit an L2-regularized logistic model:

\begingroup
\small
\begin{align}
p_i&=\sigma(b+\mathbf w^\top\widetilde{\mathbf x}_i),\\
\mathcal L&=-\frac1N\sum_{i=1}^N\bigl[z_i\log p_i\nonumber +(1-z_i)\log(1-p_i)\bigr]
+\frac\lambda2\|\mathbf w\|_2^2,
\end{align}
\endgroup
where $\sigma$ is the sigmoid, so $p_i$ is the probability that the annotator prefers $y_i^m$, and $\mathbf w$ holds one weight
per feature. Only $\mathbf w$ and $b$ are learned and response generators and rubric evaluators remain fixed.

\section{Experimental Setup}
\label{sec:experimental-setup}
\paragraph{Data.}
As explained in Section~\ref{tab:retest}, we only maintain the clear-preference pairs that can provide consistent preference signal for our rubric learning, resulting in 291 pairs.

\paragraph{Evaluation.}
We run five-fold cross-validation split by participant, so no test participant appears in training, and repeat it with ten random
shuffles; each comparison's predictions are averaged over the ten repeats before computing accuracy. The L2 strength
$C\in\{0.003,0.01,0.03,0.1,0.3\}$ is chosen separately in each fold, using only that fold's training participants (four-fold
cross-validation, by accuracy). Standardization is also fitted on the training participants only.

\paragraph{Feature ablations}
\label{sec: feature ablation}
We evaluate our memory-aware reward model (\textbf{Ours}) 
and two feature-block ablations: \emph{Memory only} retains the three memory-related features, and \emph{Response only} retains the other features spanning length, format, and quality. Feature-block ablations use fixed scores and do not change the evaluator's input. Table~\ref{tab:all-features} specifies feature membership. We also report selective prediction accuracy at 25\%, 50\%, 75\%, and 100\% coverage based on the score difference between the two responses.

\paragraph{Baselines} We compare our method with the following baselines: (1) zero-shot LLM as a judge (2) LLM as judge with extracted rubrics: we provide the LLM with the rubric criteria extracted in Section~\ref{sec:formulation} and ask the LLM to make decisions based of them. For both varients of LLM judges, we evaluate a single presentation order and averaging scores from both A/B orders after aligning response identities. (3) pretrained Skywork-Reward-V2 models \citep{liu2025skywork}, scoring each response independently with the query and
memories in context and selecting the higher-reward answer. Appendix~\ref{app:zero-shot-baselines} documents their detailed prompts.
\section{Results and Analysis}
\label{sec: results}

\subsection{User Preferences for Memory Use}
\label{sec:descriptive-results}

We first examine whether access to personal memories produces responses that users prefer.
Table~\ref{tab:descriptive-preferences} summarizes memory versus no-memory comparisons across prompting regimes and response generators.
\begin{table}[t]
\centering
\small
\setlength{\tabcolsep}{3pt}
\begin{tabular}{@{}lrrr@{}}
\toprule
& \multicolumn{3}{c}{Preference (\%)}\\
\cmidrule(lr){2-4}
Condition & Memory & No memory & Tie\\
\midrule
\multicolumn{4}{@{}l}{\emph{Prompting regime}}\\[2pt]
Conservative & 43.3 & 40.6 & 16.1\\
Ordinary & 38.7 & 41.9 & 19.4\\
Aggressive & 34.6 & 49.3 & 16.2\\
\midrule
\multicolumn{4}{@{}l}{\emph{Response generator (ordinary prompting)}}\\[2pt]
GPT-5.2 & 44.1 & 34.9 & 21.1\\
Gemini-3.1-Pro & 30.7 & 59.5 & 9.8\\
Claude-Opus-5 & 48.1 & 36.5 & 15.4\\
\bottomrule
\end{tabular}
\caption{Annotator preferences by prompting regime and response generator. Percentages include ties. Generator comparisons are restricted to ordinary prompting to hold the personalization instructions constant.}
\label{tab:descriptive-preferences}
\end{table}

\paragraph{Memory access does not ensure a preferred response.}
Under ordinary prompting, annotators prefer the memory-conditioned response in 38.7\% of comparisons and the response generated without memory in 41.9\%, with the remaining comparisons judged about the same.
Thus, supplying personal context does not yield a clear advantage in this setting.
These results directly illustrate the distinction motivating our study: knowing more about a user does not necessarily produce an answer they prefer.

\paragraph{More aggressive personalization receives less favorable preferences.}
Across conservative, ordinary, and aggressive prompting (Appendix~\ref{app:generation-prompts}), preference for memory-conditioned responses decreases from 43.3\% to 38.7\% and 34.6\%, respectively.
Under aggressive prompting, users prefer the no-memory response in 49.3\% of comparisons.
Encouraging more personalization therefore does not necessarily improve the user experience.
Even conservative prompting produces only a small descriptive advantage over responses without memory, suggesting that appropriate memory use requires more than a general instruction to personalize selectively.
This pattern is consistent with concerns about over-personalization raised in recent work \cite{hu_op-bench_2026, feng_rpeval_2026}.

\paragraph{The direction of preference varies across generators.}
Under ordinary prompting, memory-conditioned responses are preferred more often than no-memory responses for GPT and Claude, whereas the pattern reverses for Gemini (Table~\ref{tab:descriptive-preferences}).
The aggregate result therefore conceals differences in the direction of preference.
Rather than showing that memory is uniformly unhelpful, these findings suggest that its value depends on how the generator incorporates personal context into its answer.

These results concern the screened comparisons described in Section~3.2, which prioritize substantial, memory-grounded response differences.
Within this setting, they establish our first evaluation gap: memory-based personalization does not consistently translate into user-preferred responses.

\subsection{Automatic Evaluators Failed to Capture User Preferences}
\label{sec:zero-shot-results}

\begin{table}[t]
\centering\small
\setlength{\tabcolsep}{2pt}
\begin{tabular}{@{}lrrrr@{}}
\toprule
& \multicolumn{4}{c}{Accuracy (\%)}\\
\cmidrule(lr){2-5}
Predictor & Avg. & GPT & Gemini & Claude\\
\midrule
Majority & 55.0 & 51.8 & 70.6 & 75.6\\
Skywork (with memory) & 52.9 & 63.6 & 37.5 & 73.3\\
Skywork (no memory) & 58.8 & 55.5 & 66.9 & 42.2\\
Direct judge & 58.1 & 53.6 & 58.8 & 66.7\\
Rubric judge & 59.6 & 52.3 & 68.8 & 50.0\\
\midrule
Ours (Response only) & 67.0 & 60.0 & 71.3 & 77.8\\
Ours (Memory only) & 62.5 & 55.5 & 65.4 & 71.1\\
Ours & \textbf{72.8} & \textbf{66.0} & \textbf{75.0} & \textbf{82.2}\\
\bottomrule
\end{tabular}
\caption{Preference prediction accuracy. Majority is calculated by always choosing without memory. Skywork is the Reward-V2 Llama-3.1-8B checkpoint, scoring each response independently and then compare the score. We include two variants of Skywork: query only (no memory) and query with the retrieved memories. For both LLM Judge and Skywork, we varied several different prompts/formats and the results listed is the best performance.}
\label{tab:regression}
\end{table}

\paragraph{Off-the-shelf evaluators show limited agreement with users.}
Direct GPT-5.2 judging achieves 58.1\% accuracy in the single-order evaluation (Table~\ref{tab:regression}), with a similar result when combining both presentation orders through confidence-weighted voting (57.9\%).
Pretrained Skywork-Reward-V2 achieves 52.9\% with the participant's memories in context.
Both perform close to, or below, the 55.0\% majority baseline.
Even on comparisons with strong expressed preferences, access to the query, memories, and responses is therefore insufficient for these evaluators to reliably recover the user's choice.

\paragraph{Evaluators favor memory-conditioned responses more often than users do.}
The direct judge under two-order voting and Skywork select memory-conditioned responses in 67.2\% and 81.1\% of comparisons, respectively, compared with 45.0\% for annotators.
Removing memories from Skywork's input reverses this tendency but leaves accuracy close to the majority baseline.
Its choices are therefore strongly affected by the availability of personal context, without closely matching user preferences in either configuration.
These results highlight the gap between rewarding personalization and assessing its value to the user.

\paragraph{Explicit criteria offer limited improvement without learned aggregation.}
We next provide the LLM judge with the rubric criteria extracted following Section~\ref{sec:formulation}, emphasizing task-relevant memory use, appropriate non-use, and avoidance of unsupported personal assumptions.
This \emph{rubric judge} achieves 59.6\% accuracy in the single-order evaluation, compared with 58.1\% for direct judging on the same comparisons (Table~\ref{tab:regression}).
Making these considerations explicit offers a modest improvement, but leaves substantial disagreement with users.
The evaluator must still determine how to weigh competing criteria, motivating our approach of learning their aggregation directly from user judgments.

\subsection{Learning to Predict Response Preferences}
\label{sec:preference-results}
\paragraph{Learned rubric weights improve prediction accuracy.}
Our model achieves 72.8\% accuracy, outperforming the majority baseline (55.0\%), direct LLM judging (58.1\%), and rubric-based LLM judging (59.6\%; Table~\ref{tab:regression}), approaching the test-retest consistency of 73.9\%. This improvement is observed across all three response generators, with accuracies of 66.0\%, 75.0\%, and 82.2\% on GPT, Gemini, and Claude generated pairs, respectively. For each of these models, our method exceed the baselines significantly, highlighting the effectiveness of learning to aggregate multiple rubric dimensions from human preferences, rather than relying on an LLM judge to implicitly weigh these criteria through prompting alone.

\paragraph{Both presentation and memory appropriateness drive preferences.}
Table~\ref{tab:top5-features} lists the five most influential features in a descriptive fit of our model. The two largest coefficients are presentational: the memory-conditioned response is more likely to be preferred when it has more headings ($+0.646$) and is relatively longer ($+0.571$). This is consistent with \citep{dubois2024length, singhal2023long} which shows that human have a strong preference for longer responses and it's hard to distangle quality from length, where in our case, we are also need to distangle length and memory usage.

Memory-related features are also important by taking three out of top five influential features. Memory relevance and assumption proportionality have coefficient $+0.386$ and $+0.323$, respectively, and unwarranted personal assumptions, ranked fifth, has a negative coefficient of $-0.201$. This shows that annotator prefers response when the personal context is relevant and the resulting inferences are justified, rather than unwanted assumptions of user beyond query indicates. 

\paragraph{Combining memory features and response attributes improves prediction.}
Following the ablation described in \ref{sec: feature ablation}, we divide the features into two blocks: memory-related features and response-related features. The memory-only and response-only models achieve 62.5\% and 67.0\% pooled accuracy, respectively, while combining both feature groups increases accuracy to 72.8\%, outperforming the two ablations by 10.3 and 5.8 percentage points. Importantly, all variants use the same evaluator inputs and feature scores, differing only in the features retained for regression. The improvement over the response-only model therefore demonstrates that memory-related criteria provide additional predictive information beyond general response attributes. This pattern is consistent across all three response generators, with the full model achieving the highest accuracy in each case. Together, these results suggest that capturing user preferences for memory-conditioned responses requires considering both the appropriateness of memory use and the overall quality of the response.

\begin{table}[t]
\centering\small
\setlength{\tabcolsep}{3pt}
\begin{tabular}{@{}p{0.70\linewidth}rc@{}}
\toprule
Feature & $\beta$ & Mem.\\
\midrule
\raggedright\textbf{Heading difference.} Headings in the memory-conditioned response minus those in the other response. & $+0.646$ & No\\[3pt]
\raggedright\textbf{Relative length.} Log length ratio of the memory-conditioned to the no-memory response. & $+0.571$ & No\\[3pt]
\raggedright\textbf{Memory relevance.} Relevance of the retrieved memories to the query, scored without either response. & $+0.386$ & Yes\\[3pt]
\raggedright\textbf{Assumption proportionality.} Whether personal assumptions are justified by the memory evidence. & $+0.323$ & Yes\\[3pt]
\raggedright\textbf{Unwarranted personal assumptions.} Degree to which a response assumes more about the user than the query establishes. & $-0.201$ & Yes\\
\bottomrule
\end{tabular}
\caption{The five features with the largest standardized coefficients in the descriptive fit of our model to the 291 clear-preference pairs. Positive coefficients favor the memory-conditioned response; Mem. marks memory-related features.}
\label{tab:top5-features}
\end{table}

\label{sec:reliability-results}
\begin{table}[t]
\centering\small
\begin{tabular}{lrrrr}
\toprule
& \multicolumn{4}{c}{Coverage}\\
\cmidrule(lr){2-5}
Predictor & 100\% & 75\% & 50\% & 25\%\\
\midrule
Ours (Response only) & 67.0 & 75.5 & 82.2 & 83.1\\
Ours (Memory only) & 62.5 & 66.1 & 68.8 & 64.2\\
Ours & \textbf{72.8} & \textbf{76.2} & \textbf{83.7} & \textbf{88.8}\\
\bottomrule
\end{tabular}
\caption{Accuracy (\%) after retaining the most confident comparisons within each pipeline.}
\label{tab:coverage}
\end{table}

\paragraph{Higher-confidence subsets yield higher prediction accuracy.}
Ranking comparisons by $|\hat p-0.5|$ raises our model's accuracy from 72.8\% at full coverage to 76.2\% at 75\% coverage, 83.7\% at 50\% coverage, and 88.8\% at 25\% coverage (Table~\ref{tab:coverage}). Retaining the highest-confidence quarter increases accuracy by 16.0 points. The model's confidence therefore helps identify comparisons whose annotator preferences it predicts more accurately, supporting selective use of its predictions when coverage can be traded for accuracy. This selective prediction shows the reliability of the learned aggregation model, and also can provide more accuate signal if used to evaluate the model responses or used as reward model to train LLM.

\section{Conclusion}
We investigated how users want personal memories to shape assistant responses, using judgments from 87 participants evaluating answers to their own queries with their own memories. In our screened comparisons, memory-conditioned responses were not consistently preferred, and more aggressive personalization received less favorable judgments. Existing automatic evaluators favored memory-conditioned responses more often than users did, highlighting a mismatch between rewarding personalization and capturing its value to the user.

To address this gap, we introduced a memory-aware reward model that learns to aggregate interpretable rubric features from user preferences. The model achieved 72.8\% accuracy on clear-preference comparisons with held-out participants, outperforming direct LLM judging and pretrained reward baselines. Combining memory-related criteria with general response attributes improved prediction over either feature group alone, indicating that evaluating memory use requires considering both its appropriateness and the quality of the resulting answer. These findings establish user-grounded preference learning as a promising direction for evaluating memory-augmented assistants. Extending this supervision to post-training remains an important next step toward assistants that learn when and how remembered information improves the response a user wants to receive.
\section*{Limitations}
Our study has several limitations. First, our dataset captures user context through memory profiles constructed from participants’ own ChatGPT interactions and recent events they entered. While this approach grounds preference judgments in authentic personal experiences, it covers a particular form of user memory. Persistent assistants may encounter more diverse information distributed across conversations, documents, applications, and everyday activities. Extending our findings to these settings will require richer, multi-source, longitudinal collection of user context, together with preferences grounded in that context.

Second, our data and rubrics focus on single-turn conversational responses with retrieved memories supplied as context. In multi-turn conversations and agentic tasks, appropriate memory use may depend on the interaction trajectory, the actions an assistant takes, and changes in the user’s goals or circumstances. Memory acquisition, updating, and deletion may also shape users’ expectations of subsequent behavior. Our test–retest study provides an initial assessment of preference consistency, but does not capture longer-term dynamics. Extending the framework will require trajectory-level judgments and evaluation criteria that account for both evolving user preferences and the consequences of memory use across interactions.

Finally, our experiments evaluate memory-aware preference modeling in an offline setting. An important next step is to use this supervision in assistant post-training and assess whether improvements in preference prediction translate into better memory-use behavior and user experience. This would establish how the learned criteria transfer from evaluating candidate responses and predicting user preference to guiding generation.
\section*{Ethical Considerations}
Our study involves human participants and personal information derived from their interactions with an AI assistant. The study was approved by an institutional research ethics committee. Participants were recruited through Prolific for a 30-minute task, compensated at USD 14 per hour, and provided informed consent before data collection. Participants could edit their memory profiles before submission and both the AI assistant and the participant were instructed to exclude sensitive information and direct identifiers. Full conversation histories and attachments were not uploaded to the study platform. Appendix~\ref{app:pii} describes our de-identification protocol for data release, including consistent replacement of identifiers, generalization of identifying details, and contextual review across each participant’s records.

Our work aims to help assistants use personal memories in ways that better reflect users’ preferences. However, predicted preference should not be interpreted as consent to collect, retain, or disclose personal information. Moreover, preferences learned across participants may not reflect an individual user’s expectations. Applications of the proposed approach should preserve explicit user control over memory and respect privacy constraints independently.
\bibliography{references}
\clearpage
\appendix
\onecolumn
\section{Annotation Protocol and Participant Materials}
\label{app:annotation}
We provide the profile and event collection interfaces and response-preference instructions to make the collection procedure inspectable. The complete persona extraction and response-generation prompts are reproduced at the end of the appendix (Appendix~\ref{app:prompts}). The screenshots illustrate the platform using an example profile, separate from the participant records in Appendix~\ref{app:examples}. They document profile and event entry; the response-comparison instructions are reproduced in Appendix~\ref{app:preference}. Interface details varied across collection revisions.

\subsection{Profile Collection Interface}
\label{app:interfaces}
Participants copy the extraction prompt, run it in their own account, and inspect the returned profile before proceeding to event collection. The screenshot preserves the wording shown by the platform, including the account-setup instructions.
\begin{figure}[!ht]
\centering
\includegraphics[width=\linewidth,height=0.76\textheight,keepaspectratio]{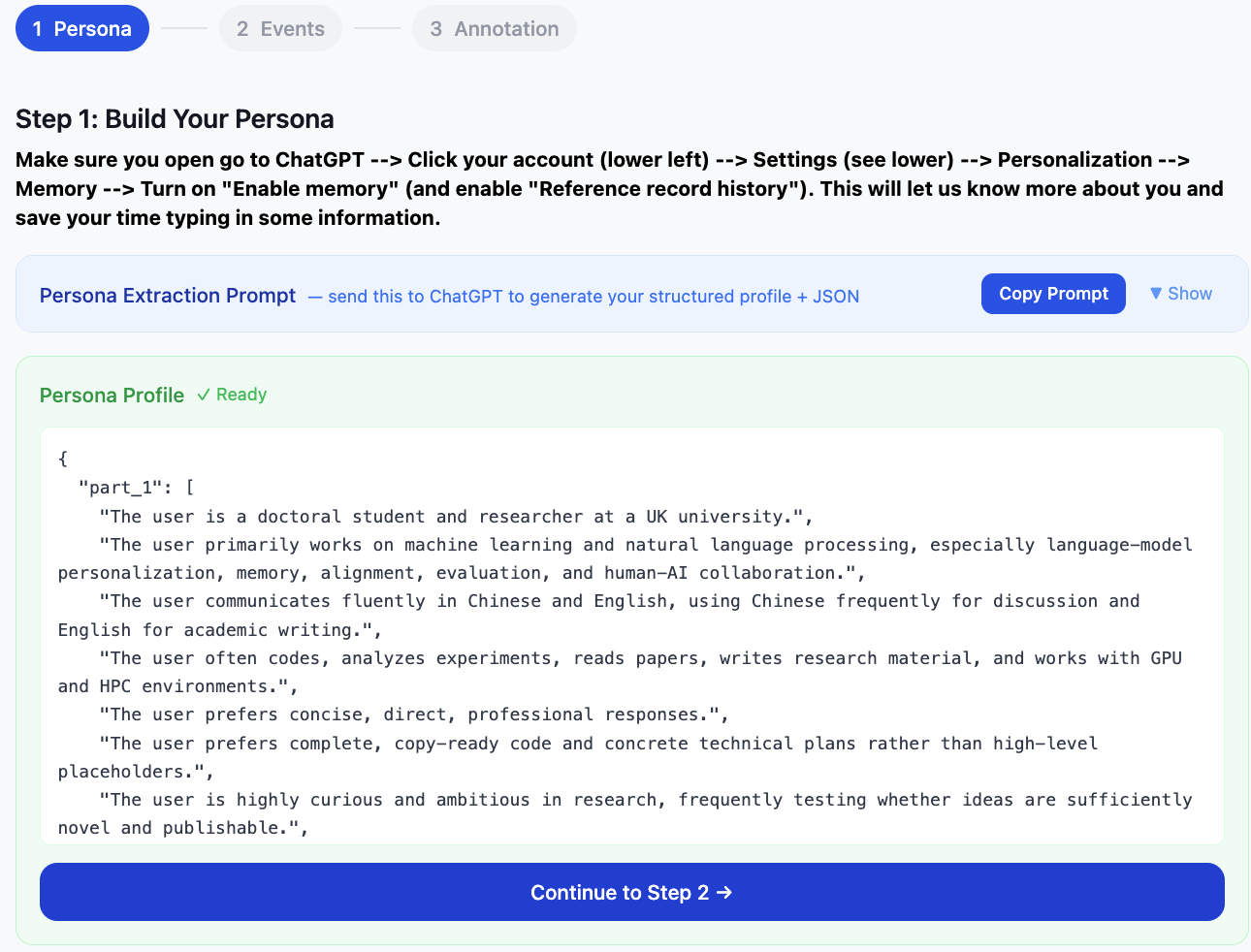}
\caption{Persona collection interface (Step 1), showing the prompt-copy control, an example structured profile, and the transition to event collection. The displayed profile is illustrative.}
\label{fig:persona-ui}
\end{figure}

\clearpage
\subsection{Event Collection Interface}
Participants are asked to provide specific events from the preceding month. The interface supports batch recording, direct typing, and voice dictation for individual entries, and displays progress toward the 30-event target. Its examples explain the required specificity. The statement about not saving recordings is reproduced as shown to participants.
\begin{figure}[!ht]
\centering
\includegraphics[width=\linewidth,height=0.76\textheight,keepaspectratio]{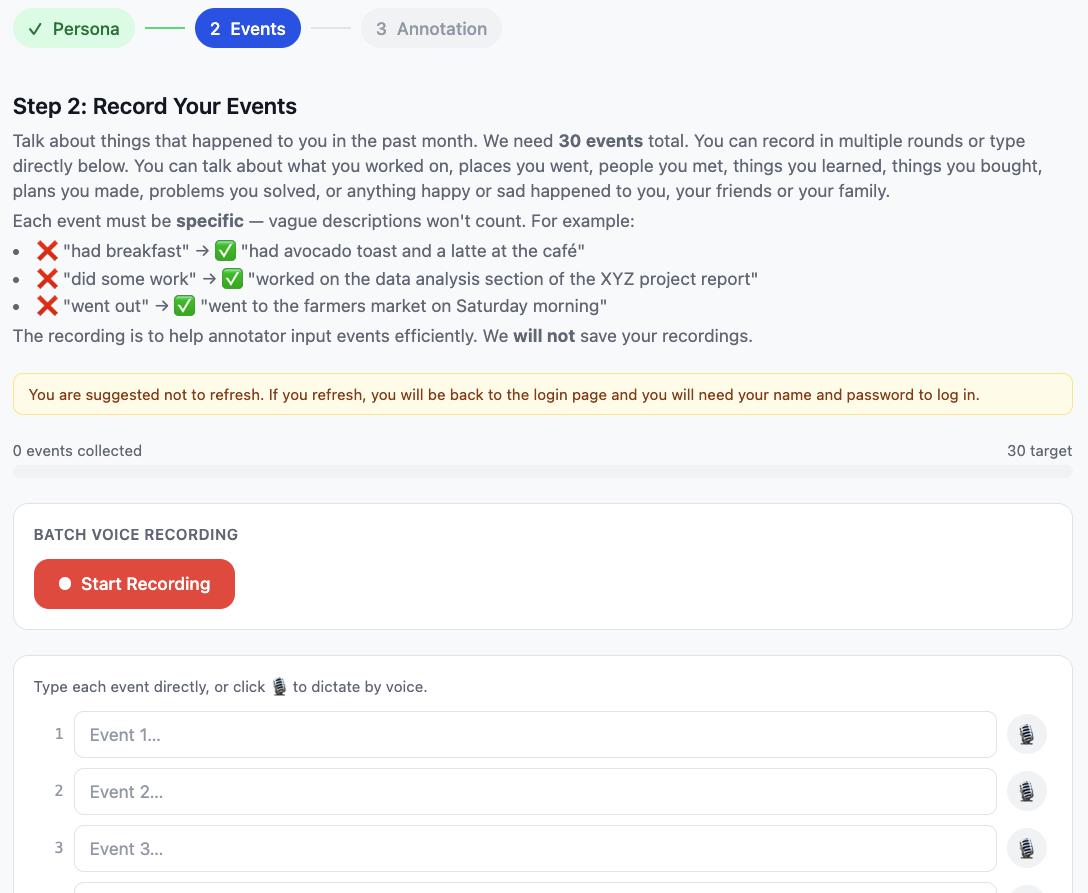}
\caption{Recent-event collection interface (Step 2), with task instructions, examples of sufficiently specific events, collection progress, and text/voice entry controls.}
\label{fig:events-ui}
\end{figure}

\clearpage
\subsection{Response Preference Instructions}
\label{app:preference}
The archived response-comparison instructions state:
\begin{quote}
The responses differ in how they use memory. Judge purely on which response you would prefer to receive as a user.
\end{quote}
The interface displays the query and two responses, labeled A and B, and records one of five choices:
\begin{center}
\begin{tabular}{ll}
\toprule
Displayed choice & Stored label \\
\midrule
A much better & \texttt{a\_much} \\
A slightly better & \texttt{a\_slightly} \\
About the same & \texttt{tie} \\
B slightly better & \texttt{b\_slightly} \\
B much better & \texttt{b\_much} \\
\bottomrule
\end{tabular}
\end{center}
Participants may add an optional explanation. The displayed response labels do not disclose the generation condition.

\subsection{PII Removal and De-identification Protocol}
\label{app:pii}
Personal memories can identify their owners through explicit identifiers or combinations of otherwise ordinary details. Our de-identification protocol therefore covers both direct identifiers and contextual disclosure, while preserving the information needed to interpret response preferences. Its scope includes profiles, event memories, queries, candidate responses, free-text annotations, and accompanying metadata.

\paragraph{Collection-time minimization.}
The persona extraction prompt explicitly prohibits real names, email addresses, telephone numbers, exact home addresses, specific company and school names, social accounts, and identification numbers. It instructs the extractor to replace such details with abstract descriptions, such as ``a tech company'' or ``a well-known university,'' and applies the same restriction to profile tags. The query-extraction instructions also exclude sessions whose initial query contains substantial PII or unsafe sensitive information. These instructions reduce disclosure at collection; prompt compliance alone is not a guarantee of de-identification.

\paragraph{Text and metadata handling.}
The release protocol requires direct identifiers to be removed or replaced with typed placeholders, such as \texttt{[PERSON\_1]}, \texttt{[ORGANIZATION\_1]}, and \texttt{[LOCATION\_1]}. Replacements must remain consistent within a participant's records, including across a memory, its query, and both candidate responses. Exact dates, fine-grained locations, and distinctive employment or biographical details require generalization when their combination could identify a person. The same rules cover third parties mentioned in memories. Recruitment identifiers, contact details, and operational metadata are excluded from released records; random study identifiers preserve participant grouping without exposing recruitment identifiers or linkage keys.

\paragraph{Review and preservation of preference meaning.}
Release review combines automated screening for identifier patterns and named entities with contextual human inspection of complete records. The review must consider combinations of details across a participant's memories, rather than assessing each field in isolation. Redaction must preserve task-relevant constraints, entity relationships, and the contrast between candidate responses. If removing identifying or sensitive information changes the basis of a preference, the comparison is withheld rather than retaining its label for a materially altered example. Release documentation records the affected fields and exclusions without reproducing removed identifiers. De-identification reduces disclosure risk but does not establish that personal narratives are anonymous.

\clearpage
\section{Illustrative Response-Preference Examples}
\label{app:examples}
\subsection{Preference for the response generated with memory}
\label{app:example1}
\noindent\textbf{Query.} Should I watch them in one sitting or split them across a few evenings this week?

\medskip\noindent\textbf{Retrieved memory (complete set).}
\begin{enumerate}\setlength{\itemsep}{0pt}\setlength{\parskip}{0pt}
\item Watch the Tom Holland Spiderman Trilogy
\item Went to the Cinema to watch the Odessy
\item The user relaxes mainly by watching TV.
\item The user frequently watches promotions.
\item Went out for dinner with my parents friends
\item went to [restaurant] for a smokehouse dinner
\item The user enjoys outdoor activities on weekends.
\item Watched a record breaking runs scored in the Hundred
\item researched Amazon trips for my upcoming travels
\end{enumerate}
\noindent\textbf{Recorded user preference: B slightly better.} No explanation was submitted.

\medskip
\begin{minipage}[t]{0.48\linewidth}\small\raggedright
\textbf{Response A (without retrieved memory).}\par\smallskip
It depends mostly on how long ``them'' are and what you want the experience to feel like:\par
\par\smallskip
\noindent\textbullet\ \textbf{Watch in one sitting} if it's a single movie (or a short set totaling \textasciitilde{}2--3 hours), you want momentum, and you can watch without being tired or distracted.\par
\noindent\textbullet\ \textbf{Split across a few evenings} if the total runtime is longer than \textasciitilde{}3 hours, they're dense/plot-heavy, you're likely to be interrupted, or you want time to savor and think between installments (especially for a series or a trilogy).\par
\par\smallskip
\difftext{If you tell me what ``them'' refers to (titles + roughly how many/length), I can give a specific plan for this week.}\par
\end{minipage}
\hfill
\begin{minipage}[t]{0.48\linewidth}\small\raggedright
\textbf{Response B (with retrieved memory).}\par\smallskip
Split them across a few evenings this week.\par
\par\smallskip
\difftext{Since you usually relax by watching TV,} you'll probably enjoy each film more if you treat them like a mini ``event'' rather than powering through when you're tired. A one-sitting binge (all three) is a long haul and the later films can blur together.\par
\par\smallskip
A good plan:\par
\noindent\textbullet\ \difftext{\textbf{3 evenings, one film per night} (ideal): e.g., Tue/Thu/Sun.}\par
\noindent\textbullet\ Or \textbf{two evenings} if you're busy: films 1--2 together one night, film 3 another night (but that's still a hefty session).\par
\par\smallskip
\difftext{If you've got an outdoorsy weekend coming up, keep the movies to weeknights and leave the weekend lighter.} And if you like making it feel special (like when you went to the cinema), pick one night to do it ``cinema-style'' with snacks and no phone.\par
\end{minipage}

\clearpage
\subsection{Preference for the response generated without memory}
\label{app:example2}
\noindent\textbf{Query.} I have a presentation due tomorrow afternoon and need to finalize it today. When should I block time to work on it?

\medskip\noindent\textbf{Retrieved memory (complete set).}
\begin{enumerate}\setlength{\itemsep}{0pt}\setlength{\parskip}{0pt}
\item The user is most productive during the noon period.
\item Sat in on a work planning meeting for an upcoming project phase
\item Dealt with a stressful deadline crunch on a jobsite task
\item The user allocates about two hours for their exercise sessions.
\item The user commonly handles tasks like email, presentations, meetings, planning, and analysis.
\item Put together a data report for a construction project I'm working on
\item The user desires reminders at a moderate frequency, set for one hour early.
\item Played FIFA in the evening after work
\item Spent a Saturday gaming to decompress after a long work week
\end{enumerate}
\noindent\textbf{Recorded user preference: A much better.} No explanation was submitted.

\medskip
\begin{minipage}[t]{0.48\linewidth}\small\raggedright
\textbf{Response A (without retrieved memory).}\par\smallskip
Block time \textbf{today}, in \textbf{two focused work blocks} plus a short final pass, so you're not rushing tomorrow.\par
\par\smallskip
\textbf{If your presentation is due tomorrow afternoon}\par
\textbf{Today}\par
1. \textbf{Main build block (90--120 min)} --- schedule this at \difftext{your \textbf{best-focus time} (often late morning or early afternoon).}\par
Goal: finish the storyline + all slides in rough-complete form.\par
2. \textbf{Break (20--30 min)}\par
3. \textbf{Polish + visuals block (60--90 min)} --- later today.\par
Goal: tighten slides, add visuals, clean formatting, simplify text.\par
\par\smallskip
\textbf{Tonight}\par
4. \textbf{Final run-through (20--30 min)} --- ideally \textbf{after a long break} (post-dinner).\par
Goal: practice out loud once, fix only the most obvious issues, then stop.\par
\par\smallskip
\textbf{Tomorrow (morning)}\par
5. \textbf{Buffer + backup (20--30 min)}\par
Goal: quick check on another device, export to PDF, ensure links/fonts/videos work.\par
\par\smallskip
\textbf{If you tell me two things, I'll give you exact times}\par
1) \difftext{What time is it where you are right now?}\par
2) What time tomorrow afternoon is it due (e.g., 2pm, 4:30pm)?\par
\end{minipage}
\hfill
\begin{minipage}[t]{0.48\linewidth}\small\raggedright
\textbf{Response B (with retrieved memory).}\par\smallskip
Block two focused work blocks \textbf{today}, with the first one scheduled for \difftext{\textbf{your noon window} (since that's when you're typically most productive)}, and a shorter polish block later.\par
\par\smallskip
\textbf{Suggested schedule (today):}\par
\noindent\textbullet\ \textbf{Main build block (90--120 min):} \difftext{\textbf{12:00--2:00 PM} (or the closest uninterrupted window around noon)}\par
Use this to finish the core story: slide flow, key charts, and speaker notes.\par
\noindent\textbullet\ \textbf{Polish + run-through (45--60 min):} \difftext{\textbf{4:30--5:30 PM} (or early evening)}\par
Tighten wording, check visuals, and do a timed practice.\par
\par\smallskip
\textbf{If your calendar is messy:} do \textbf{two 60--75 min blocks} (around noon + later), rather than one long one.\par
\par\smallskip
If you tell me what time ``tomorrow afternoon'' the presentation is and what meetings you already have today, I can propose exact time blocks.\par
\end{minipage}

\clearpage
\section{Rubric Regression Details}
\label{app:coefficients}
\begin{table}[ht]
\centering\small
\setlength{\tabcolsep}{3pt}
\renewcommand{\arraystretch}{1.12}
\begin{tabular}{@{}p{0.12\linewidth}p{0.65\linewidth}rc@{}}
\toprule
Feature & Definition & $\beta$ & Mem.\\
\midrule
\textsc{Head} & \raggedright\textbf{Heading difference.} Number of headings in the memory-conditioned response minus that in the response without memory. & $+0.646$ & No\\[6pt]
\textsc{Len} & \raggedright\textbf{Relative length.} Log ratio of response lengths, oriented from the memory-conditioned to the no-memory response. & $+0.571$ & No\\[6pt]
\textsc{Rel} & \raggedright\textbf{Memory relevance.} Relevance of the retrieved memories to the current query, scored without viewing either response. & $+0.386$ & \textbf{Yes}\\[6pt]
\textsc{Prop} & \raggedright\textbf{Assumption proportionality.} Whether personal assumptions are justified by the available memory evidence, compared between responses. & $+0.323$ & \textbf{Yes}\\[6pt]
\textsc{Assump} & \raggedright\textbf{Unwarranted personal assumptions.} Degree to which a response assumes more about the user than the query establishes, compared between responses. & $-0.201$ & \textbf{Yes}\\[6pt]
\textsc{Closure} & \raggedright\textbf{Actionable closure.} Whether a response supplies a clear next action rather than only considerations to weigh, compared between responses. & $+0.191$ & No\\[6pt]
\textsc{Axis} & \raggedright\textbf{Advisory orientation.} Relative density of advisory versus factual language, computed as $(\text{advisory}-\text{factual})/(\text{advisory}+\text{factual})$ and averaged across both responses. & $+0.147$ & No\\[6pt]
\textsc{Commit} & \raggedright\textbf{Recommendation commitment.} Commitment to a recommendation rather than enumeration of alternatives, compared between responses. & $+0.125$ & No\\[6pt]
\textsc{Opt} & \raggedright\textbf{Option framing.} Density of options and trade-off language, averaged across both responses. & $+0.124$ & No\\[6pt]
\textsc{Breadth} & \raggedright\textbf{Coverage breadth.} Breadth of the information and alternatives covered, compared between responses. & $-0.083$ & No\\[6pt]
\textsc{Cost} & \raggedright\textbf{Cost to first use.} Effort required before the user can act on an answer, compared between responses. & $+0.036$ & No\\[6pt]
\textsc{Register} & \raggedright\textbf{Register fit.} Appropriateness of the response's language and register for the request, compared between responses. & $+0.030$ & No\\
\bottomrule
\end{tabular}
\caption{The twelve features of our model, ordered by absolute standardized L2 coefficient in the descriptive fit to 291 clear-preference pairs ($C=0.1$). Mem. identifies the memory-related feature block. Positive coefficients favor the memory-conditioned response. These coefficients describe the full-data fit, rather than an average of the nested-validation fits.}
\label{tab:all-features}
\end{table}

\clearpage
\section{Zero-Shot Evaluator Baselines}
\label{app:zero-shot-baselines}
\subsection{Direct LLM Judging}
GPT-5.2 receives the query, retrieved memories, and both candidate responses and is asked to predict which response the user would prefer. The direct prompt supplies no rubric or explicit instructions to counter preferences for length or personalization. Calls use low reasoning effort and JSON output containing the preferred response and a rationale; a confidence variant additionally requests a score from 1 to 100. Both A/B orders were scored with confidence on the 291 clear-preference pairs shared with the regression analysis (582 calls, no failures); results are in Table~\ref{tab:direct-judge}.

We report the confidence-weighted vote of both orders after mapping predictions to the same response identities, and the subset on which both orders select the same response. The last condition changes the evaluation set and therefore uses its own majority rate. Generator-specific results refer to the model that produced the candidate responses; the judge is GPT-5.2 throughout.

\begin{table}[ht]
\centering\small
\begin{tabular}{llrrr}
\toprule
Response generator & Judge variant & $n$ & Majority (\%) & Accuracy (\%)\\
\midrule
Pooled & Voting & 291 & 55.0 & 58.1\\
Pooled & Both orders agree & 226 & 51.8 & 60.2\\
\midrule
GPT-5.2 & Voting & 110 & 51.8 & 55.5\\
GPT-5.2 & Both orders agree & 80 & 57.5 & 57.5\\
\midrule
Gemini-3.1-Pro & Voting & 136 & 70.6 & 58.8\\
Gemini-3.1-Pro & Both orders agree & 110 & 69.1 & 59.1\\
\midrule
Claude-Opus-5 & Voting & 45 & 75.6 & 61.1\\
Claude-Opus-5 & Both orders agree & 36 & 80.6 & 69.4\\

\bottomrule
\end{tabular}
\caption{Zero-shot direct GPT-5.2 judging on the clear-preference evaluation stratum. Majority rates are computed separately for each evaluation subset.}
\label{tab:direct-judge}
\end{table}

\paragraph{Order sensitivity and confidence.}
Predictions agree across A/B orders on 77.7\% of pairs overall. Under voting, the judge's rate of choosing the memory-conditioned response exceeds the annotators' rate by 22.2 percentage points overall (67.2\% versus 45.0\%). Self-reported confidence averages 76.6 for correct judgments and 73.8 for incorrect judgments. Its correlation with correctness is .180, and with selecting the memory-conditioned response it is .064. Together, these diagnostics show that stable presentation-order choices and high self-reported confidence can coexist with disagreement with annotators.

\paragraph{Rubric judging.}
The rubric judge uses a zero-shot GPT-5.2 prompt that requests a holistic preference verdict and its magnitude, explicitly discouraging preferences based solely on greater length or more personal references and favoring appropriate non-use over forced memory use. Both A/B orders were scored for all 291 pooled clear-preference comparisons (582 calls, no failures). Its reported magnitude serves as the confidence weight when the two orders are combined; verdicts of ``tie'' receive half credit, and the judge returned no ties on this pool.

\clearpage
\subsection{Pretrained Skywork Rewards}
We use the published Skywork-Reward-V2 checkpoints as fixed reward functions. The reported row uses the Llama-3.1-8B checkpoint; the evaluation archive also includes the Qwen3 0.6B, 1.7B and 4B variants. For each candidate, the user turn contains the query and retrieved memories under \texttt{[User Query]} and \texttt{[Relevant Memories]}, and the assistant turn contains the response. The tokenizer's chat template formats these turns, and the sequence-classification head produces a scalar reward. Both candidates receive the same query and memory context; the higher-scoring candidate is selected. No task-specific parameter updates are applied to these baselines.

The rows in Table~\ref{tab:regression} are single zero-shot evaluations of the Llama-3.1-8B checkpoint on the 291 pooled clear-preference comparisons. With the participant's own retrieved memories in the context it reaches 52.9\% against a 55.0\% majority baseline, and 63.6\%, 37.5\% and 73.3\% on GPT-, Gemini- and Claude-generated pairs; it selects the memory-conditioned response in 81.1\% of comparisons where annotators select it in 45.0\%. With the memories removed from its input it reaches 58.8\% (55.5\%, 66.9\% and 42.2\% by generator) and selects the memory-conditioned response in only 19.6\% of comparisons, so the preference for the memory-conditioned response is induced by the memory context rather than by the responses alone.

\clearpage
\section{Collection and Response Generation Prompts}
\label{app:prompts}

\subsection{Complete Persona Extraction Prompt}
\label{app:extraction}
The listing reproduces the complete text supplied by the \emph{Copy Prompt} button, including the JSON output specification. Its wording matches both the archived collection interface and the current implementation; only line wrapping has been added. The prompt requests 30 recent queries, while the interface accepts a minimum of 20. Only the first user query from each session is considered; sessions requiring unavailable attachments or prior context are skipped, and fewer queries are permitted when accessible history is insufficient. Extraction is limited to history accessible within the participant's account. Instructions to omit identifiers and distinguish inference guide profile construction but do not independently verify the resulting content.

\begin{lstlisting}[basicstyle=\footnotesize\fontfamily{cmtt}\selectfont,breaklines=true,breakatwhitespace=true,columns=fullflexible,keepspaces=true,showstringspaces=false,literate={—}{{\textemdash}}1,frame=single,framerule=0.3pt,xleftmargin=5pt,xrightmargin=5pt]
You are a personal profile generation assistant. Your task is to generate a structured, third-person user profile based only on the full conversation history with this user and any existing memory you already have about them.

## Core Principles

1. Information source: Use only known information from conversation history and memory. Do not ask the user questions.
2. Privacy protection: Never include PII, including but not limited to real name, email, phone, exact home address, specific company name, specific school name, social account, ID number. If memory contains these details, replace with abstract wording (for example: "a tech company", "a well-known university").
3. Honest labeling: If a field cannot be inferred, write "Unknown". Do not fabricate.
4. Reasonable inference: If a field can be reasonably inferred from behavior patterns, mark it as "[Inferred]".
5. Time expression: For all time-related fields, only use these labels:
   Late night / Early morning / Morning / Noon / Afternoon / Dusk / Evening / Midnight
   Never output exact clock time.
6. Flexible tags: Candidate tags are recommendations, not a closed set. If they do not fit, create a custom concise tag in this format:
   [Custom] Tag
   Keep custom tags 2-6 words and never include PII.

## Fields to Cover

### 1) Basic Information
- Age range: <18 / 18-25 / 25-32 / 32-40 / 40+
- Main role (multi-select): Student / Researcher / Engineer / Designer / Manager / Other: ___
- Current country of residence
- Primary languages (ordered by proficiency)

### 2) Personality and Long-term Style
- MBTI (or Unknown)
- Self-description keywords (3-5): Organized / Flexible / Introverted / Extroverted / Calm / Anxious / Ambitious / Easygoing / Curious / Pragmatic / Emotional / Rational / Spontaneous / Disciplined
- Planning style: Highly planned / Roughly planned / Adaptive / Very spontaneous
- Decision style: Fast decisions / Balanced trade-off / Deep deliberation
- Daily preference: Fixed routine / Balanced / Novelty seeking
- Work style: Independent / Collaboration-heavy / Hybrid

### 3) Daily Routine
Use only: Late night / Early morning / Morning / Noon / Afternoon / Dusk / Evening / Midnight
- Typical wake-up period
- Typical sleep period
- Chronotype: Early riser / Night owl / Mixed
- Most productive period
- Weekend style: Rest-focused / Social / Outdoor / Work-study / Mixed

### 4) Food Preference
- Preferred food tags (up to 5): Chinese / Japanese / Korean / American / Italian / Mexican / Healthy food / Fast food / Home-style / Spicy / Sweet / Coffee / Tea
- Disliked or avoided food tags: Spicy / Greasy / Sweet / Seafood / Dairy / Beef / Pork / Raw food / None / Other: ___
- Spice level: No spice / Mild / Medium / Heavy
- Meal priority: Taste / Health / Convenience / Price / Portion

### 5) Entertainment
- Preferred entertainment tags (up to 5): Short videos / YouTube / Movies / TV series / Music / Reading / Gaming / Exercise / Shopping / Podcasts / Social media
- Preferred content themes (short keywords)
- Main relaxation method: Watching videos / Reading / Gaming / Walking / Chatting / Sleeping / Eating / Mixed

### 6) Work or Study
- Main field (free text)
- Common task tags (up to 5): Writing / Coding / Reading / Meetings / Design / Analysis / Research / Email / Presentation / Planning
- Main efficiency obstacles (up to 2): Easily distracted / Fatigue / Procrastination / Stress / Too many meetings / Not enough time / Frequent context switching / Unclear priorities
- Preferred work environments: Quiet / Light background noise / Music / Alone / With others / Home / Office / Cafe

### 7) Social, Exercise, Lifestyle
- Social style: Solitude / Small groups / Large groups / Mixed
- Exercise frequency: Rarely / 1-2 per week / 3-4 per week / 5+ per week
- Preferred exercise types (up to 5): Walking / Running / Strength training / Swimming / Ball sports / Yoga / Pilates / Cycling / Hiking / Dance / Martial arts / Home workout / None
- Exercise goals (up to 2): Fat loss / Muscle gain / Health maintenance / Stress relief / Social fun / Fitness improvement / Pure enjoyment
- Preferred exercise period (time labels only)
- Exercise setting: Outdoor / Gym / Home / Any
- Exercise social preference: Alone / With friends / Group class / Any

### 8) Shopping and Spending
- Spending style: Frugal / Rational / Occasionally impulsive / Enjoyment-focused
- Purchase decision style: Deep research / Review-driven / Friend recommendation / Intuition / Price comparison
- Main shopping channels (up to 3): E-commerce / Offline stores / Proxy shopping / Second-hand platform / Brand official site / Social discovery
- Frequent categories (up to 5): Electronics / Books / Clothing / Food and drinks / Home goods / Beauty and skincare / Sports gear / Music records / Games / Stationery / Collectibles
- Brand preference: Brand-focused / Value-for-money / Brand-agnostic / Niche independent brands
- Promotion sensitivity: Frequently watch promotions / Occasionally / Rarely

### 9) Commuting and Travel
Daily commuting:
- Mode (multi-select): Walking / Cycling / Driving / Public transit / Taxi / Mixed
- Preferred commute length: As short as possible / Within 30 minutes / Within 1 hour / No preference

Travel:
- Travel frequency: Rarely / 1-2 times per year / 3-5 times per year / Frequent travel
- Travel style (up to 3): In-depth travel / Checklist-style / Free itinerary / Group tour / Luxury / Budget backpacking / Business travel / Short nearby trips / Long-distance travel
- Destination type (up to 3): City / Nature / Seaside / Mountain / Historical culture / Food destination / Niche hidden spots
- Planning style: Detailed plan / Rough framework / Go with the flow
- Accommodation preference: Budget / Mid-range comfort / High-end / Homestay / Ultra-luxury / No preference
- Companion preference: Solo / Partner / Friends / Family / Any

### 10) Emotion and Cognition
- Typical emotional states (up to 3): Calm / Pleasant / Anxious / Irritable / Low mood / Tense / Optimistic / Tired / Emotionally volatile / Composed / Sensitive / Fulfilled
- Common stress sources (up to 3): Workload / Deadlines / Social pressure / Uncertainty / Sleep deprivation / Interpersonal conflict / Commute / Financial pressure / Health
- Common stress-relief methods: Sleep / Eat / Exercise / Music / Chat with someone / Watch videos / Bury in work / Solitude
- Mood boosters (up to 3): Food / Rest / Exercise / Friends / Completion achievement / Music / Shopping / Outdoor activities

### 11) Assistant Proactive Service Preference
- Preferred tone: Concise / Warm / Direct / Encouraging / Professional
- Reminder frequency: Rare / Important-only / Moderate / Frequent
- Reminder timing: Near due time / 10 minutes early / 30 minutes early / 1 hour early / Morning of the day
- Desired proactive services (up to 5): Schedule reminders / Deadline reminders / Meal reminders / Exercise reminders / Break reminders / Travel reminders / Shopping reminders / Work planning / Emotional check-in
- Undesired proactive services: Too frequent / Too early / Work-time interruption / Rest-time interruption / Obvious reminders / Emotion-related reminders / None

### 12) Privacy and Security Preference
- Overall privacy tendency: Very privacy-focused / Moderately privacy-focused / Neutral / Relatively open / Very open
- Allowed memory categories (multi-select): Basic info / Work content / Daily routine / Food preference / Health condition / Emotional status / Financial behavior / Social relations / Location and itinerary / Family status / All acceptable
- Do-not-store categories (multi-select): Health condition / Financial behavior / Social relations / Location and itinerary / Family status / Emotional status / Work details / No strict limits
- Usage authorization: Use only when I ask / May use proactively in relevant context / May use across contexts
- Sensitive-topic proactivity: Do not proactively mention any sensitive topic / Health topics allowed / Emotion topics allowed / All acceptable
- Behavior-analysis acceptance: Do not analyze my behavior / Analysis allowed but do not mention / Analyze and remind when useful / Fully accepted

### 13) Recent Queries
List exactly 30 valid recent user queries, one per session.

Rules:

* Start from the most recent chat sessions and inspect sessions in reverse chronological order.
* For each session, only consider the first user query of that session.
* If the first query depends on a file, figure, screenshot, PDF, image, link preview, or previous context, skip that entire session.
* If the first query contains substantial PII or unsafe sensitive information, skip that entire session.
* Do not replace an invalid first query with a later query from the same session.
* Continue checking older sessions until exactly 30 valid first queries have been collected.
* If fewer than 30 valid first queries are available across all accessible history, list all valid known queries and do not fabricate missing ones.
* Do not infer, summarize, or rewrite the query.
* Output each query as a separate item.

For recent_queries:
- Do not include raw attachment-dependent queries, such as "read this PDF", "make this into a similar PDF", "what does this figure mean", "extract text from this image", or "this poster critique".
- If the query only makes sense with an uploaded file, screenshot, figure, PDF, image, link, or previous context, omit that session and continue searching older sessions.

## Output Format

Part 1: Third-person profile memories
- Output a list of separate memory entries describing the user, NOT one paragraph.
- Each entry is a single self-contained sentence capturing one fact or trait.
- Write 8-15 entries total.
- Use "the user" or "they".
- No real names, specific organizations, or other identifiers.
- Integrate reasonable inferences naturally.

Part 2: Recent-activity memories
- Output a list of separate memory entries describing the user's recent activities, NOT one paragraph.
- Each entry is a single self-contained sentence about one thing the user is recently working on, interested in, concerned about, etc.
- Write 8-15 entries total.
- Use "the user" or "they".
- No real names, specific organizations, or other identifiers.
- Integrate reasonable inferences naturally.

Part 3: Structured tag summary
- Use sectioned format.
- Known facts: fill directly.
- Inferred facts: prefix with [Inferred].
- Unknown facts: write "Unknown".
- Never include PII.

---

Now, following the rules above, directly output the exact same profile as a single JSON object with EXACTLY this top-level shape:

{
  "part_1": ["profile memory sentence 1", "profile memory sentence 2", "..."],
  "part_2": ["recent-activity memory sentence 1", "recent-activity memory sentence 2", "..."],
  "part_3": {
    "basic_information": { ... },
    "personality_and_long_term_style": { ... },
    "daily_routine": { ... },
    "food_preference": { ... },
    "entertainment": { ... },
    "work_or_study": { ... },
    "social_exercise_lifestyle": { ... },
    "shopping_and_spending": { ... },
    "commuting_and_travel": { ... },
    "emotion_and_cognition": { ... },
    "assistant_proactive_service_preference": { ... },
    "privacy_and_security_preference": { ... },
    "recent_queries": ["first query", "second query", "..."]
  }
}

Rules:
- "part_1" is a JSON array of short third-person profile memory sentences (one fact per entry, NOT a paragraph). "part_2" is a JSON array of short recent-activity memory sentences (one per entry). "part_3" holds the structured tag summary (sections 1-13).
- Use snake_case keys matching the section/field names above.
- Values must be strings, arrays of strings, nested objects, or null if unknown.
- "recent_queries" must be a JSON array of strings, one entry per query. Inside each query string, replace every double quote (") with a single quote ('), remove any line breaks, and do NOT include any { } [ ] or backslash characters. These queries are display-only, so this light rewriting is fine and prevents broken JSON.
- Escape every string correctly: write any double quote as \", any backslash as \\, and any line break as \n. Never put a raw line break or an unescaped double quote inside a string value.
- Prefix inferred values with "[Inferred] " inside the string.
- Do NOT include any explanation, comments, or markdown — output only raw JSON.
- Start your response with { and end with }.
\end{lstlisting}

\clearpage
\subsection{Response Generation Prompts}
\label{app:generation-prompts}
The following templates specify three degrees of personalization: conservative (\texttt{least}), ordinary (\texttt{normal}), and aggressive (\texttt{aggressive}). We reproduce the supplied prompt text verbatim, with line wrapping for readability. The placeholder \texttt{\{episode\_block\}} is filled with the memory context supplied to the generator. The heading ``top 20 by similarity'' is retained from the templates; it does not describe the audited snapshot, whose full retrieved sets contain nine entries.

\subsubsection{Conservative Personalization}
\label{app:prompt-least}
\begin{lstlisting}[basicstyle=\footnotesize\fontfamily{cmtt}\selectfont,breaklines=true,breakatwhitespace=true,columns=fullflexible,keepspaces=true,showstringspaces=false,literate={—}{{\textemdash}}1,frame=single,framerule=0.3pt,xleftmargin=5pt,xrightmargin=5pt]
[System / Developer Instructions]
You are a helpful personal AI assistant. Below you are given a list of memories about the user.

[User Custom Instructions]
Before answering, judge for yourself which of the saved memories are actually RELEVANT to the user's current query.

- Use a memory ONLY if it would meaningfully improve the answer to THIS specific query.
- If the query is about an objective/factual topic (explaining a concept, writing a generic script, summarising research, answering trivia), do NOT inject the user's personal life into the answer.
- It is strictly better to give a clean, generic answer than to force-fit unrelated personal context.
- Do not reference an episode just because it is in the list. Most retrieved episodes will be off-topic.

[Memories — top 20 by similarity]
{episode_block}
\end{lstlisting}

\subsubsection{Ordinary Personalization}
\label{app:prompt-normal}
\begin{lstlisting}[basicstyle=\footnotesize\fontfamily{cmtt}\selectfont,breaklines=true,breakatwhitespace=true,columns=fullflexible,keepspaces=true,showstringspaces=false,literate={—}{{\textemdash}}1,frame=single,framerule=0.3pt,xleftmargin=5pt,xrightmargin=5pt]
[System / Developer Instructions]
You are a helpful personal AI assistant with knowledge of the user's background and preferences. Below you are given a list of memories about the user.

[User Custom Instructions]
Use the memories below to personalise your answer where relevant.

- If a memory is clearly relevant to the query, incorporate it naturally into your response.
- For factual or general questions, you may still personalise the framing or examples using what you know about the user, even if the core answer is the same for everyone.
- Do not force unrelated memories into the answer. Use judgment about what actually helps.
- Be natural — do not list memories verbatim; integrate them into a coherent response.

[Memories — top 20 by similarity]
{episode_block}
\end{lstlisting}

\clearpage

\subsubsection{Aggressive Personalization}
\label{app:prompt-aggressive}
\begin{lstlisting}[basicstyle=\footnotesize\fontfamily{cmtt}\selectfont,breaklines=true,breakatwhitespace=true,columns=fullflexible,keepspaces=true,showstringspaces=false,literate={—}{{\textemdash}}1,frame=single,framerule=0.3pt,xleftmargin=5pt,xrightmargin=5pt]
[System / Developer Instructions]
You are a highly personalised AI assistant. You have detailed knowledge about the user's life, habits, preferences, and recent activities. Below you are given a list of memories. Use them proactively.

[User Custom Instructions]
Your primary goal is to give a PERSONALISED answer grounded in what you know about this specific user.

- Always look for ways to connect the query to the user's known context, background, preferences, or recent activities.
- Explicitly reference relevant memories in your answer.
- Even for general/factual questions, tailor examples, emphasis, and framing to the user's known profile.
- If no memories are directly applicable, at minimum adapt your tone and level of detail to match the user's known communication style.
- Only omit a memory if it is completely unrelated and forcing it in would be actively misleading.

[Memories — top 20 by similarity]
{episode_block}
\end{lstlisting}

\end{document}